\documentclass[10pt,twocolumn,letterpaper]{article}

\usepackage{iccv}
\usepackage{times}
\usepackage{epsfig}
\usepackage{graphicx}
\usepackage{amsmath}
\usepackage{amssymb}
\usepackage{multirow}
\usepackage{caption}
\usepackage{subcaption}
\usepackage{array}

\usepackage[breaklinks=true,bookmarks=false]{hyperref}

\iccvfinalcopy 

\def\iccvPaperID{9105} 

\ificcvfinal\fi

\begin{document}

\title{Error Estimation for \\Single-Image Human Body Mesh Reconstruction}

\author{Hamoon Jafarian\\
Ontario Tech University\\
Oshawa, Ontario, Canada\\
{\tt\small hamoon.jafarian@ontariotechu.ca}
\and
Faisal Z. Qureshi\\
Ontario Tech University\\
Oshawa, Ontario, Canada\\
{\tt\small faisal.qureshi@ontariotechu.ca}
}

\maketitle
\ificcvfinal\thispagestyle{empty}\fi

\begin{abstract}
Human pose and shape estimation methods continue to suffer in situations where one or more parts of the body are occluded.  More importantly, these methods cannot express when their predicted pose is incorrect. This has serious consequences when these methods are used in human-robot interaction scenarios, where we need methods that can evaluate their predictions and flag situations where they might be wrong.  This work studies this problem.  We propose a method that combines information from OpenPose and SPIN---two popular human pose and shape estimation methods---to highlight regions on the predicted mesh that are least reliable.  We have evaluated the proposed approach on 3DPW, 3DOH,  and Human3.6M datasets, and the results demonstrate our model's effectiveness in identifying inaccurate regions of the human body mesh. Our code is available at \url{https://github.com/Hamoon1987/meshConfidence}.
\end{abstract}

\section{Introduction}
\label{sec:intro}

\begin{figure}
    \setlength{\tabcolsep}{1pt} 
  \centering
  \begin{tabular}{c |c |c c}
        Input & SPIN & \multicolumn{2}{c}{Our method} \\
        \includegraphics[width=0.21\linewidth]{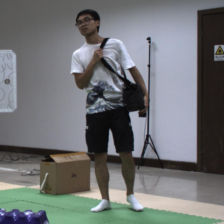} & 
        \includegraphics[width=0.21\linewidth]{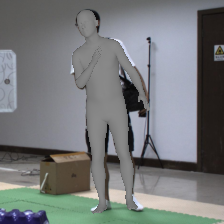} & 
        \includegraphics[width=0.21\linewidth]{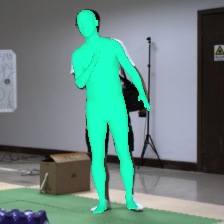} & 
        \includegraphics[width=0.21\linewidth]{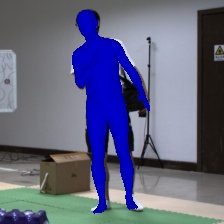} \\
        \includegraphics[width=0.21\linewidth]{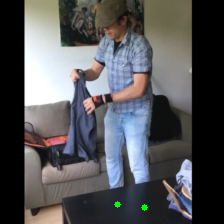} & 
        \includegraphics[width=0.21\linewidth]{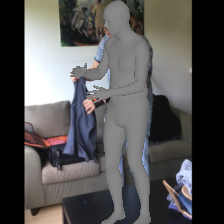} & 
        \includegraphics[width=0.21\linewidth]{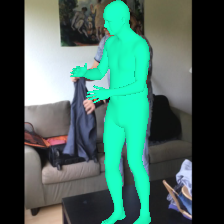} & 
        \includegraphics[width=0.21\linewidth]{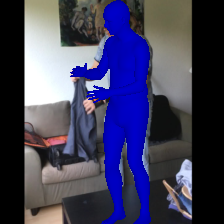} \\
        \includegraphics[width=0.21\linewidth]{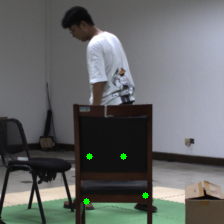} & 
        \includegraphics[width=0.21\linewidth]{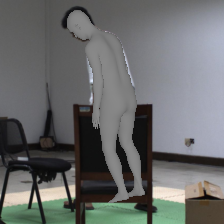} & 
        \includegraphics[width=0.21\linewidth]{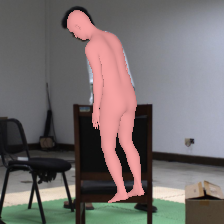} & 
        \includegraphics[width=0.21\linewidth]{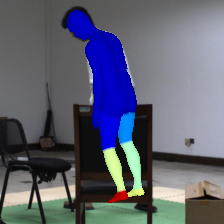} \\
        \includegraphics[width=0.21\linewidth]{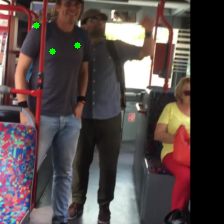} & 
        \includegraphics[width=0.21\linewidth]{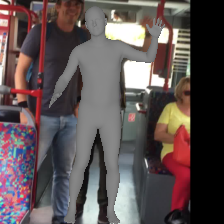} & 
        \includegraphics[width=0.21\linewidth]{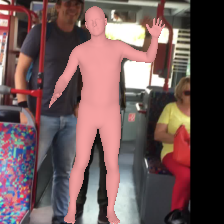} & 
        \includegraphics[width=0.21\linewidth]{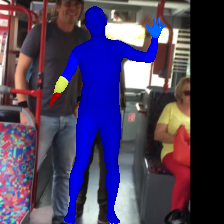}\\
        \includegraphics[width=0.21\linewidth]{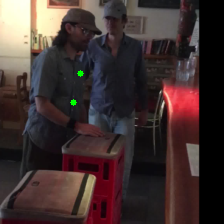} & 
        \includegraphics[width=0.21\linewidth]{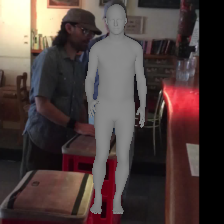} & 
        \includegraphics[width=0.21\linewidth]{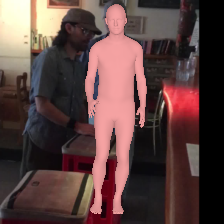} & 
        \includegraphics[width=0.21\linewidth]{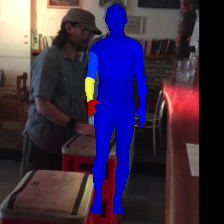}\\
        \includegraphics[width=0.21\linewidth]{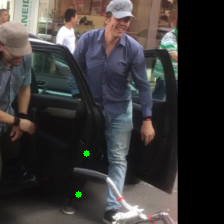} & 
        \includegraphics[width=0.21\linewidth]{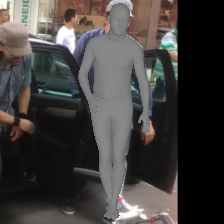} & 
        \includegraphics[width=0.21\linewidth]{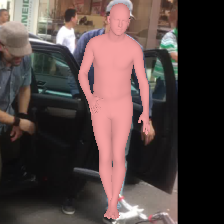} & 
        \includegraphics[width=0.21\linewidth]{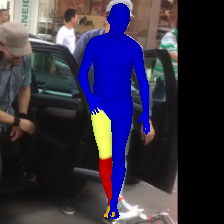} \\       
  \end{tabular}
    \caption{Human body mesh recovery using the SPIN model vs. our method.  The input images are in the first column; the recovered meshes using SPIN are shown in the second column.  Our method identifies whether or not the recovered mesh corresponds to a partially occluded human and highlights regions of the mesh that are unreliable since these represent parts of the human body that are not visible in the image.}
    \label{fig:other-methods}
\end{figure}

The applications of Human Body Mesh Recovery (HMR) are diverse and numerous. HMR, for example, is useful for Human-Robot Interaction (HRI) scenarios, where accurate 3D mesh representation is essential for ensuring safe interactions~\cite{martinez2018real}. Drones, self-driving cars, and human-robot collaborative manufacturing systems are some examples where a three-dimensional understanding of the environment and humans is critical for reliable operation~\cite{liu2017human}. Additionally, the animation and movie industries can benefit significantly from HMR by simplifying the process of character motion capture (MOCAP) and reducing the costs involved~\cite{tung2017self}. Other areas such as part and foreground segmentation, computer-assisted coaching, and virtual try-on can also leverage the capabilities of 3D mesh recovery to enhance their outcomes~\cite{tian2022recovering}.

The task of estimating a human body mesh from a single RGB image is an active area of research that has garnered significant interest in the field of computer vision.  Kolotoures and colleagues~\cite{kolotouros2019learning} proposed SPIN that achieves impressive results on single image human body mesh recovery.  SPIN represents a significant improvement in human pose and shape estimation over prior methods, and it is now a widely adopted baseline in the field.  A number of recent methods attempt to recover human body mesh in the presence of occlusions~\cite{zhang2020object,kocabas2021pare,khirodkar2022occluded}.  None of these methods, however, provide a confidence score for the recovered mesh.  The ability to tell whether or not the recovered mesh is correct or to identify parts of the mesh that may be inaccurate is particularly relevant for human-robot interaction scenarios.  A robot, for example, can choose to halt its operation if it deems that the recovered mesh is not reliable.  Alternatively a robot may adjust its viewpoint to achieve a better reconstruction if it identifies one or more parts of the mesh to be unreliable.

\begin{figure}
    \setlength{\tabcolsep}{1pt} 
  \centering
  \begin{tabular}{c c c}
        \includegraphics[width=0.32\linewidth]{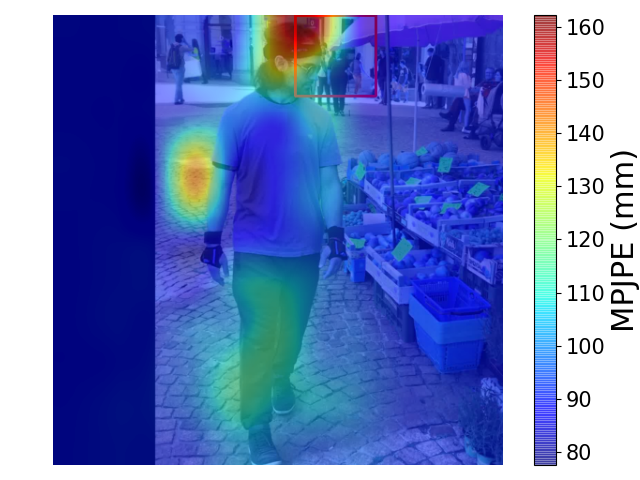} &
        \includegraphics[width=0.32\linewidth]{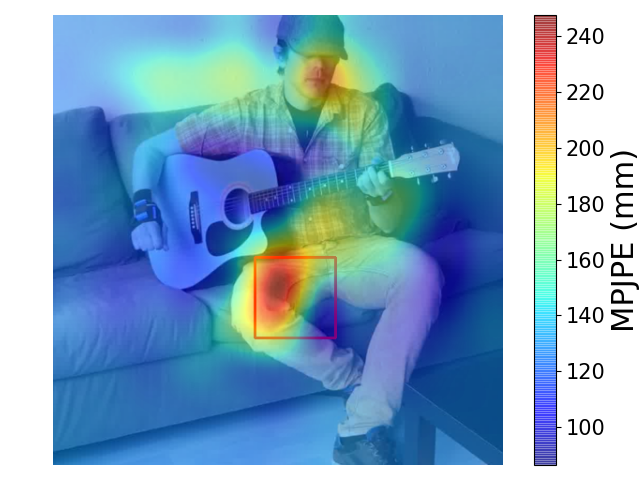} &
        \includegraphics[width=0.32\linewidth]{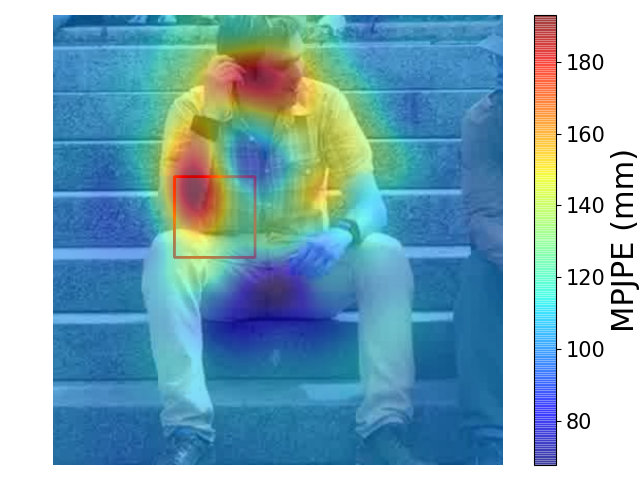} \\
        \includegraphics[width=0.32\linewidth]{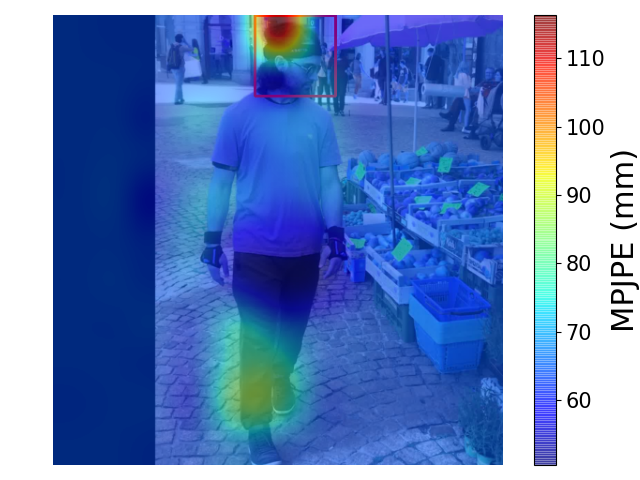} & 
        \includegraphics[width=0.32\linewidth]{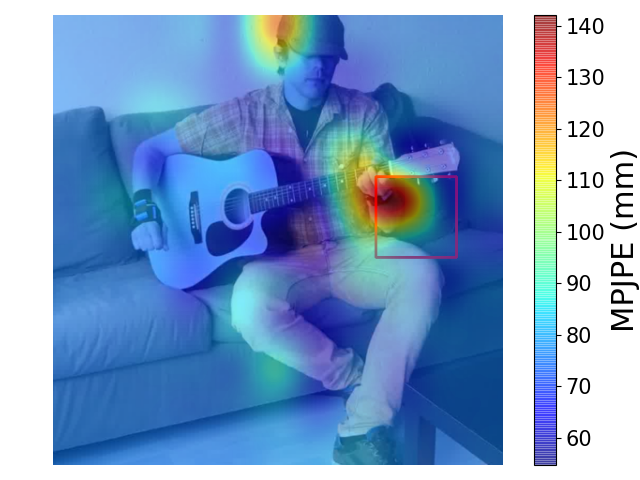} & 
        \includegraphics[width=0.32\linewidth]{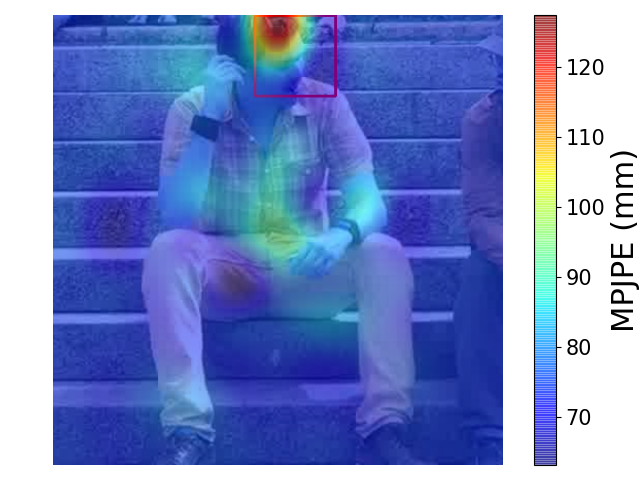} \\
  \end{tabular}
    \caption{Pixel Locations based Occlusion Sensitivity Analysis.  A $40\times40$ square (gray) occluder is moved across the image and MPJPE values are computed (for each) for SPIN (top-row) and OpenPose (bottom-row) models.  The heatmaps highlight locations in the image that strongly affects the performance of the two models.  Both models are sensitive to occlusions in regions shown in red. Image size is $224\times224$ and the stride is selected to be $20$.}
    \label{fig:sensitivityAnalysis_image}
\end{figure}

Here we tackle the problem of estimating the error in the reconstructed human body meshes.  We propose a method that fuses information from SPIN and OpenPose~\cite{cao2017realtime} to highlight regions of the recovered mesh that {\it may} be inaccurate (Figure~\ref{fig:other-methods}).  OpenPose estimates human joints' keypoints, and it is able to identify joints that are not visible in the image.  The proposed method leverages the observation that SPIN and OpenPose agree when the person is visible in the image; whereas, these two methods disagree when the person is partially occluded.  We have used {\it sensitivity analysis} to quantify the disagreement between SPIN and OpenPose models under occluded settings.  The differences between the joints' keypoints estimated by OpenPose and those constructed by projecting the human body mesh recovered by SPIN are fed into two multi-layer perceptron networks to compute an error estimate for each region of the mesh.  In Figure~\ref{fig:other-methods} (last column from right) regions shown in red depict mesh parts with the lowest reliability.  Note that these regions correspond to the parts of the human body that are not visible in the image.

To the best of our knowledge, this work represents the first attempt at estimating error in single-image 3D human body mesh reconstructions.  The contributions of the work presented here are: 1) location-based and joint-based occlusion sensitivity analysis to quantify the relationship between the disagreement of OpenPose and SPIN joint location estimates and the ``true'' error; 2) a mesh classifier that identifies whether or not the recovered mesh is reliable; and 3) a worst joint classifier that selects the least reliable joint.  This work represents a significant step towards improving the safety and reliability of those human-robot interactions that rely upon accurate reconstructions of human body mesh by providing additional information about the confidence and reliability of the estimated mesh.

\begin{figure}
    \setlength{\tabcolsep}{1pt} 
  \centering
  \begin{tabular}{c c |c c}
        \multicolumn{2}{c}{\small 3DPW} & \multicolumn{2}{c}{\small H36M} \\
        \tiny{\textcolor{blue}{min: 101} - \textcolor{red}{max: 130}} &
        \tiny{\textcolor{blue}{min: 111} - \textcolor{red}{max: 141}} &
        \tiny{\textcolor{blue}{min: 69} - \textcolor{red}{max: 96}} &
        \tiny{\textcolor{blue}{min: 113} - \textcolor{red}{max: 141}} \\
        \includegraphics[width=0.23\linewidth]{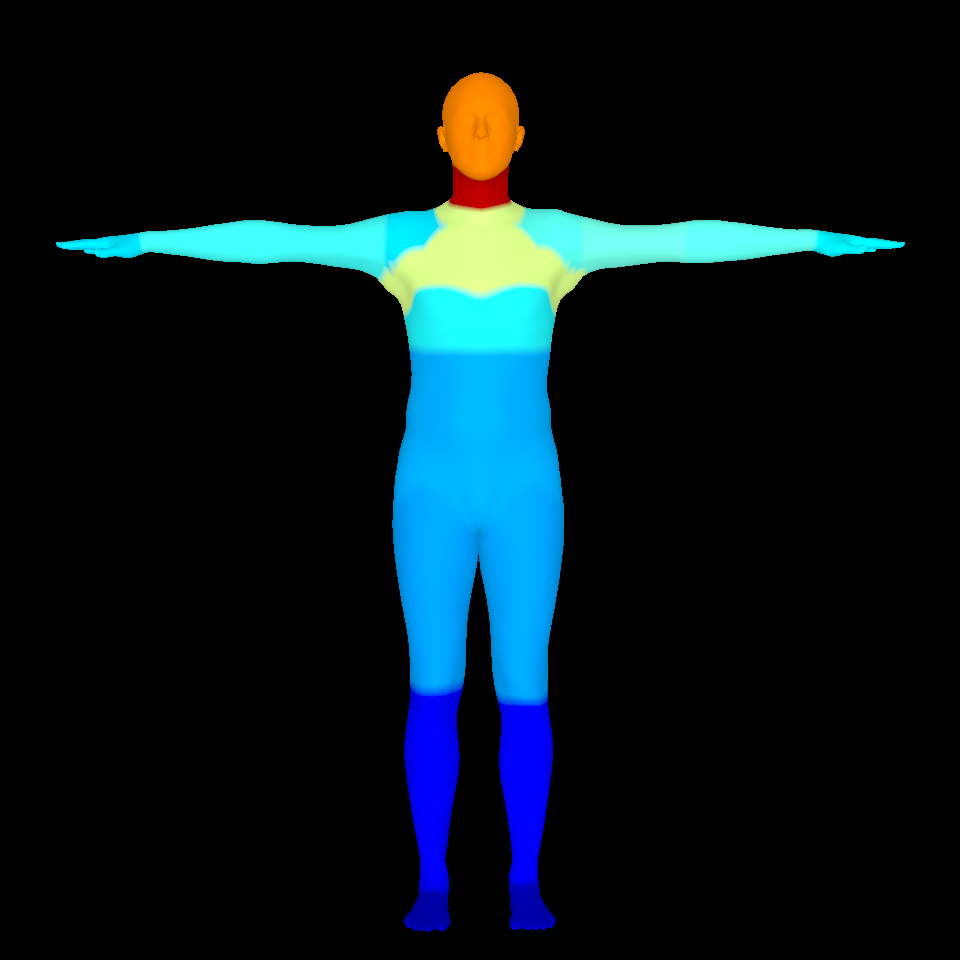} &
        \includegraphics[width=0.23\linewidth]{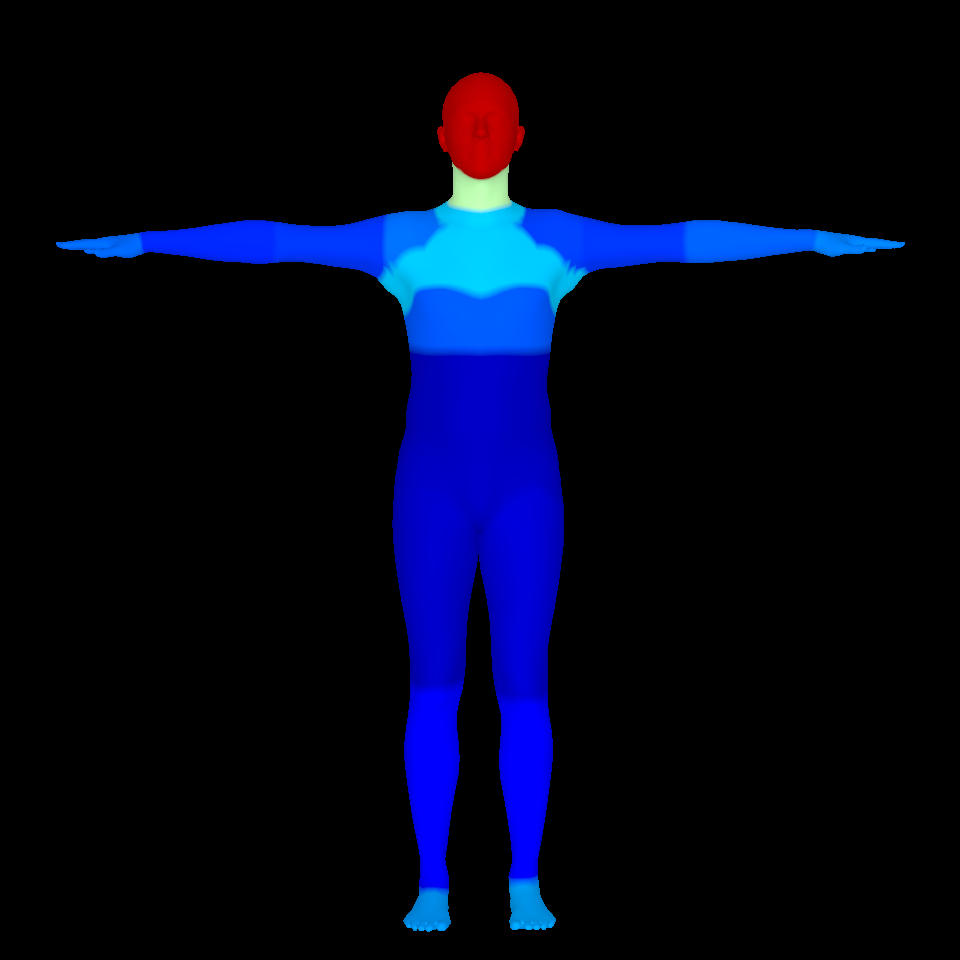} &
        \includegraphics[width=0.23\linewidth]{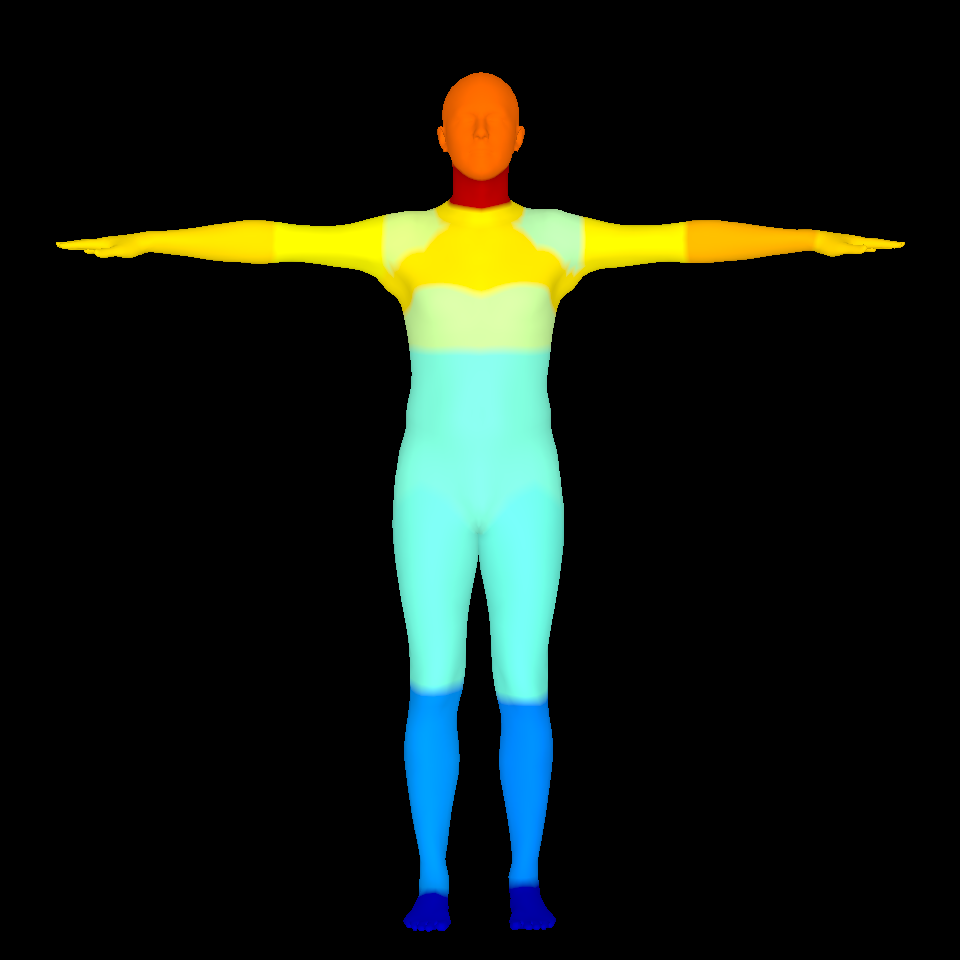} & 
        \includegraphics[width=0.23\linewidth]{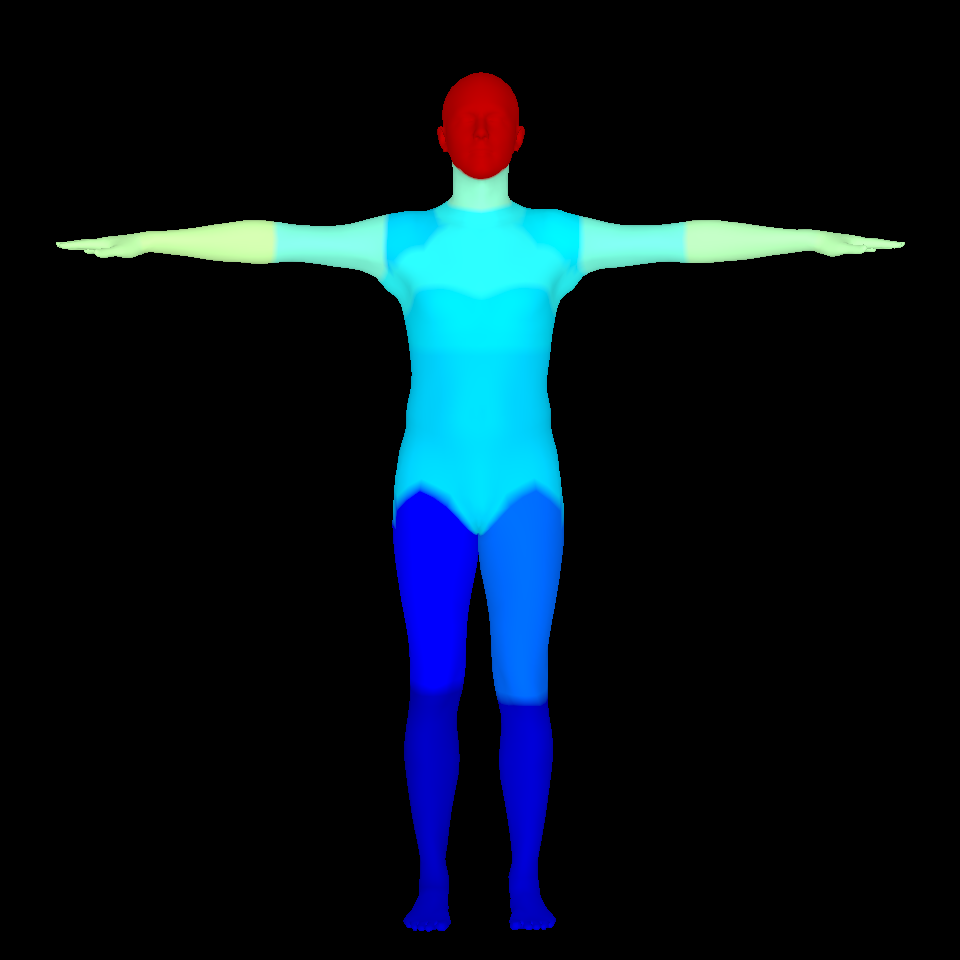}  \\
  \end{tabular}
    \caption{Joints based Occlusion Sensitivity Analysis.  For every image in 3DPW and H36M datasets, a square occluder is pasted over each joint in turn and MPJPE values are computed for SPIN (left) and OpenPose (right) models.  MPJPE errors for each joint are visualized by highlighting the vertices (of the mesh) that correspond to each joint.  The figures depict model performance if one of the joints is occluded.  This figure is best viewed in color.}
    \label{fig:sensitivityAnalysis}
\end{figure}

\begin{figure*}
     \centering
     \includegraphics[width=.9\textwidth]{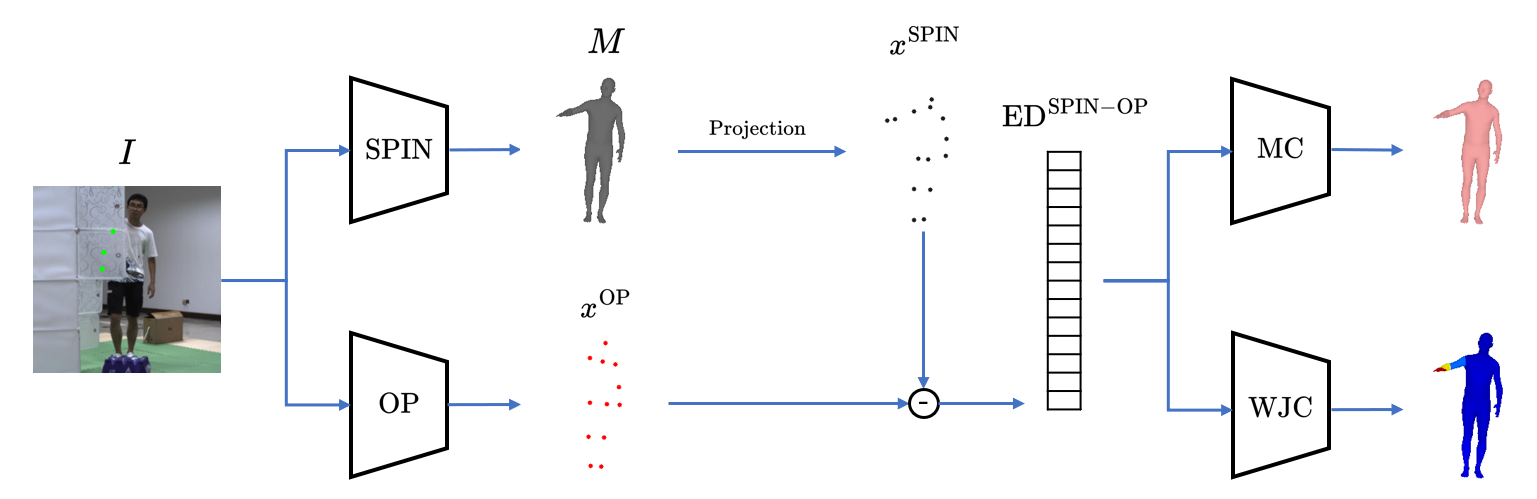}
     \caption{Overview of the Proposed Framework. The input image $I$ is passed through the SPIN and OpenPose models. Then, the estimated SPIN mesh ($M$) is regressed and projected into 2D joint coordinates. Comparing the results with the OpenPose predicted 2D joint positions, Estimation Difference (ED) is obtained. Afterward, ED is employed to train the Mesh Classifier (MC) and Worst Joint Classifier (WJC) that decide the SPIN mesh quality and detect the least reliable parts of the mesh, respectively.}
     \label{fig:framework}
\end{figure*}
\section{Related Work}
\label{sec:rel}

\textbf{2D Keypoint Estimation.} 2D keypoint estimation aims to localize body joints within an image.  Joints' keypoint estimation comes in two flavours: regression-based methods~\cite{sun2017compositional, luvizon2019human, zhang2020distribution} and detection-based methods~\cite{sun2017human, chen2017adversarial}.  Top-down approaches achieve better accuracy in multi-person scenarios; however, bottom-up approaches are often faster and are better suited for real-world applications that demand real-time performance~\cite{gamra2021review}.  Pishcgulin et al.~\cite{pishchulin2016deepcut} proposed DeepCut, a CNN-based body part detector, and Insafutdinov et al. \cite{insafutdinov2016deepercut} improved the DeepCut model by employing a ResNet-based deep part detector.  Cao et al. \cite{cao2017realtime} introduced Part Affinity Fields (PAFs) that encode the position and orientation of human body parts and propose OpenPoase, an accurate, fast, and robust model for multi-person joints' keypoints estimation.

\textbf{3D Pose and Shape Estimation.} Broadly speaking 3D pose and shape estimation methods are divided into two classes: optimization-based methods that deform a canonical pose to match the image~\cite{bogo2016keep, zanfir2018monocular, lassner2017unite} and regression-based methods that directly estimate the mesh from the image~\cite{kanazawa2018end, omran2018neural, pavlakos2018learning}.  Optimization-based methods achieve good results; however, these are slow and require careful initialization.  The regression-based methods, on the other hand, are difficult to train to achieve high-quality meshes~\cite{tian2022recovering}.  Kolotoures et al. \cite{kolotouros2019learning} proposed SPIN method for human body mesh reconstruction that employs optimization to provide explicit 3D supervision to train a regressor to construct high-quality meshes.  Hybrid models achieve state-of-the art performance.  These benefit from the 3D pose and shape estimation models' ability to capture the realistic body structure and combine it with the higher accuracy of keypoint estimation models~\cite{li2021hybrik}.

\textbf{Occlusion Handling.} Inspired by random erasing~\cite{zhong2020random} and synthetic occlusion~\cite{dvornik2018modeling} techniques exploited in classification and object detection tasks, some researchers suggest that data augmentation could be a suitable solution against occlusion. In this scenario, the images are occluded throughout the training process, and the model is taught to perform better against occlusion~\cite{sarandi2018robust, ke2018multi}. Others modified the model architecture to improve the model's robustness against occlusion. Zhang et al.~\cite{zhang2020object} uses a partial UV map model to convert the occluded human body to an image inpainting problem. Georgakis et al.~\cite{georgakis2020hierarchical} develop a prior-informed regressor that knows the hierarchical structure of the human body, and the experiments show that this method improves the model performance against occluded cases. Kocabas et al.~\cite{kocabas2021pare} implemented the soft attention mechanism for the HMR problem, resulting in a considerable improvement of the model's robustness against occlusion. The developed part attention regressor (PARE) learns to rely on visible body parts to reason about the occluded parts.

\begin{figure*}
    \setlength{\tabcolsep}{1pt} 
  \centering
  \begin{tabular}{c c}
        \rotatebox{90}{\hspace{0.6cm}Regular} & 
        \includegraphics[width=0.95\linewidth]{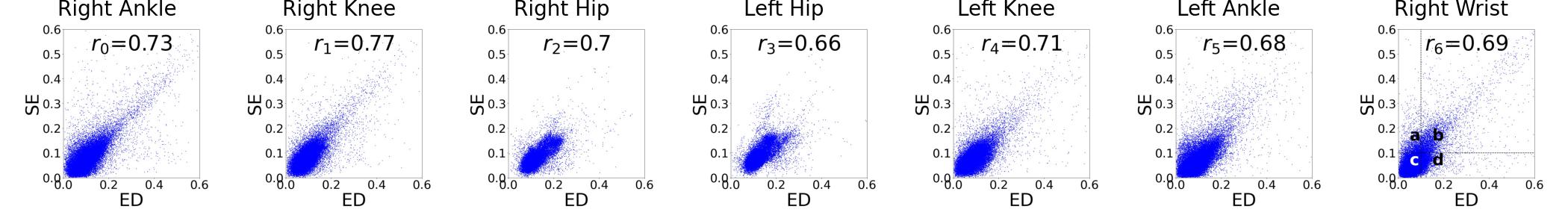} \\ 
        \rotatebox{90}{\hspace{0.6 cm}Occluded} &
        \includegraphics[width=0.95\linewidth]{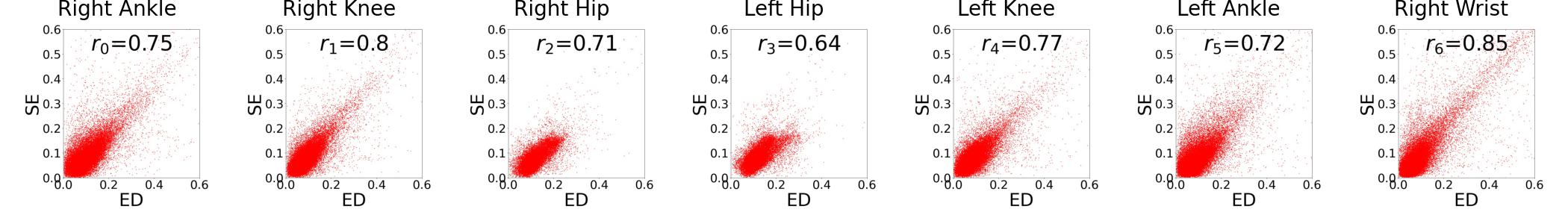} \\ 
        \rotatebox{90}{\hspace{0.6cm}Regular} & 
        \includegraphics[width=0.95\linewidth]{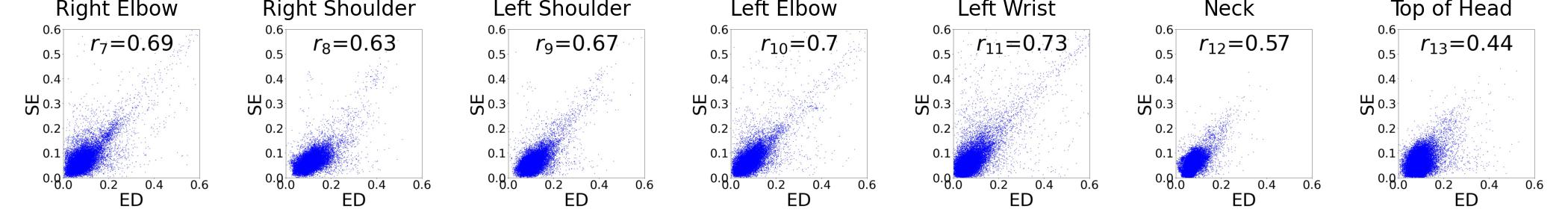} \\ 
        \rotatebox{90}{\hspace{0.6 cm}Occluded} &
        \includegraphics[width=0.95\linewidth]{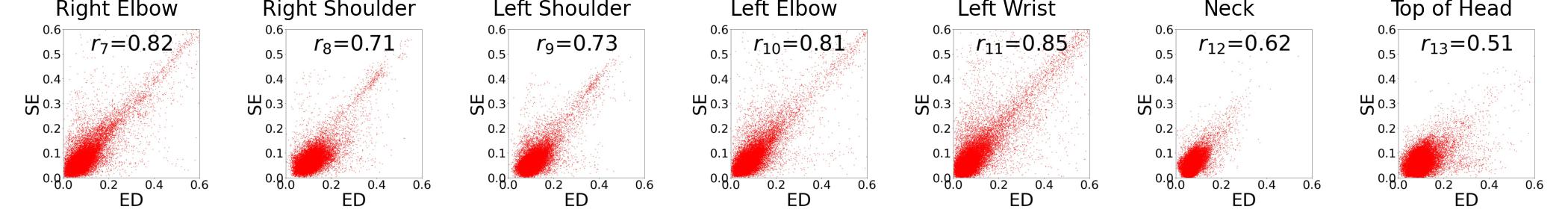} \\ 
  \end{tabular}
    \caption{Pearson Correlation Coefficient. We calculate the correlation coefficient for each joint throughout the 3DPW dataset.  The average value in the absence of occlusions is $\overline{r}=0.67$.  This value jumps to $0.735$ for the occluded version of the 3DPW dataset.  These values suggest a positive correlation between $\text{ED}$ and $\text{SE}$.  Four regions (a, b, c and d) are indicated in the top-right plot.  Region a denotes False Positive scenarios, i.e., the estimated joint location is inaccurate, however, the proposed model has failed to identify it.  Region d denotes False Negative scenarios where the estimated joint location is erroneously labelled inaccurate.  Combining information from multiple joints helps deal with these scenarios.}
    \label{fig:correlation}
\end{figure*}
\section{Occlusion Sensitivity Analysis}

We experiment with two approaches to visualize and understand the effects of partial occlusions of the human body on the performance of SPIN and OpenPose models.  The first approach captures the sensitivity (of both methods) to occluded regions for a given image.  The second approach, on the other hand, shows the sensitivity to an occluded joint over the entire dataset.

The first approach is inspired by~\cite{zeiler2014visualizing, kocabas2021pare}, where a square occluder is pasted onto different pixel locations in the image.  Both the size and the stride of the occluder can be changed.  Similar to~\cite{zeiler2014visualizing}, we use a grey colored square.  The occluded images are passed to SPIN and OpenPose models and the errors are recorded. The performance of both models is measured using the Mean per Joint Position Error (MPJPE) that is defined as the mean value of the Euclidean distance between the ground truth and the predicted locations of all the joints.  SPIN model recovers an SMPL mesh $M$.  Using a pre-trained regressor $W$, it is possible to estimate 3D joint locations $X = W M$, where $X \in \mathtt{R}^{K \times 3}$.  $K = 14$ refers to the number of joints.  MPJPE for SPIN model is
\begin{equation}
\text{MPJPE}^{\text{SPIN}}_{(m,n)} = \underset{k}{\text{Mean}} \  \| X_{(m,n)} - X^{\text{gt}}\|.
\end{equation}
Here $X_{(m,n)}$ denotes 3D joint locations when occluder is centered at location $(m,n)$.  $X^{\text{gt}}$ denotes ground truth 3D joint locations.
For the OpenPose model, which estimates 2D joint locations $x \in \mathtt{R}^{K \times 2}$,
\begin{equation}
\text{MPJPE}^{\text{OP}}_{(m,n)} = \underset{k}{\text{Mean}} \ \| x_{(m,n)} - x^{\text{gt}}\|,
\end{equation}
where $x_{(m,n)}$ and $x^{\text{gt}}$ are 2d joint estimates when occluder is centered at $(m,n)$ and ground truth 2d locations, respectively.
Figure~\ref{fig:sensitivityAnalysis_image} plots MPJPE scores for both models using a heatmap.  The figure shows how partial occlusions affect the performance of the two methods as measured by MPJPE.

For the second approach, the square occluder is used to hide specific joints through the entire dataset.  Where as the first approach captures the occlusions sensitivity to particular image locations, the second approach finds occlusions sensitivity to different joints.  In this case
\begin{equation}
\text{MPJPE}^{\text{SPIN}}_{k} = \underset{i}{\text{Mean}} \  \underset{k}{\text{Mean}}\  \| X_{i,k} - X_i^{\text{gt}}\|,
\end{equation}
where $i$ indices over images, $k$ indices over images, $X_{i,k}$ denotes 3D joints' locations estimations for image $i$ when occluder is centered on joint $k$.  $X_i^{\text{gt}}$ is ground truth 3D joint locations for image $i$.  Similarly,
\begin{equation}
\text{MPJPE}^{\text{OP}}_{k} = \underset{i}{\text{Mean}} \ \underset{k}{\text{Mean}} \ \| x_{i,k} - x_i^{\text{gt}}\|.
\end{equation}
Here $x_{i,k}$ refers to OpenPose joint estimates for image $i$ when the occluder is centered at joint $k$ and $x_i^{\text{gt}}$ denotes ground truth 2D joints for image $i$.  Figure~\ref{fig:sensitivityAnalysis} visualizes MPJPE values for both methods on an SMPL mesh.  Every vertex of a joint is associated with one or more joints, and each vertex is assigned a color using $\text{MPJPE}_k$ values, where $k$ belongs to the set of joints associated with this vertex.  These colors visualize the sensitivity of the two methods to an occluded joint.

\section{Method}\label{method}

Figure~\ref{fig:framework} illustrates the proposed method for assigning an error estimate to different regions of the reconstructed human body mesh.  It comprises three steps: 1) SPIN model is used to estimate ``2D'' joint locations, 2) OpenPose model is used to recover 2D joint locations, 3) The difference between the 2D joint estimates for SPIN and OpenPose is used to assign a confidence score to the mesh.  When SPIN and OpenPose models correctly estimate a joint position, the estimated coordinates are close to each other and adjacent to the ground truth. However, based on the sensitivity analysis, when the models' estimated positions are inaccurate, we expect the joint position estimations to be dissimilar. Hence, the distance between the models' outputs 
\begin{equation}
  \text{ED}_i = \|x_{i}^{\text{SPIN}} - x^{\text{OP}}_i\|,
  \label{eq:conf}
\end{equation}
can be considered as a proxy for the confidence in the recovered human body mesh.  Here $x_i^{\text{OP}}$ are 2D joint estimates for OpenPose and $x_i^{\text{SPIN}}$ are \emph{projected} 2D joint estimates for SPIN.  $\text{ED}_i \in \mathtt{R}^{K}$ and $i$ refers to the image.

\begin{figure}
    \setlength{\tabcolsep}{1pt} 
  \centering
  \begin{tabular}{c c | c c}
        Input & Output & Input & Output \\
        \includegraphics[width=0.23\linewidth]{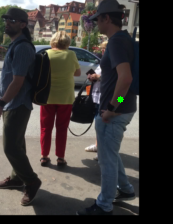} & 
        \includegraphics[width=0.23\linewidth]{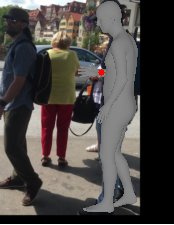} & 
        \includegraphics[width=0.23\linewidth]{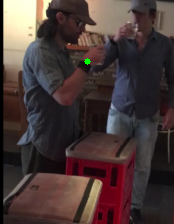} & 
        \includegraphics[width=0.23\linewidth]{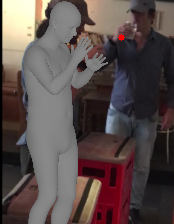} \\
  \end{tabular}
    \caption{The right wrist is occluded in the first input image, making both Openpose and SPIN models misestimate the right wrist's position.  However, these wrong estimations are adjacent. The green dot shows the ground truth position, and the red dot represents the OpenPose estimation of the right wrist. In the second case, OpenPose is confused by the other person's right wrist and makes a wrong estimation, while the SPIN model accurately estimates the right wrist. These are two samples that negatively affect the correlation between ED and SE.}
    \label{fig:failCases}
\end{figure}

To investigate the hypothesis that ED is a useful proxy for the confidence in the recovered mesh, we calculate the correlation between the ED and the SPIN model’s error
\begin{equation}
  \text{SE}_i = \|x_{i}^{\text{SPIN}} - x^{\text{gt}}_i\|.
  \label{eq:SE}
\end{equation}
The Pearson correlation coefficient of joint $k$ which is shown by $r_k$ is calculated using
\begin{equation}
    r_k = \text{Corr}([\text{ED}_{0,k}, ...,\text{ED}_{n,k}], [\text{SE}_{0,k}, ...,\text{SE}_{n,k}]),
    \label{eq:correlation}
\end{equation}
where $n$ stands for the number of images in the dataset. 
Since the OpenPose model provides 2D estimates, it can only be compared to the 2D projection of the SPIN model output. Hence, SE only captures the 2D error of the SPIN model. Additionally, the OpePose model does not provide any estimations for the undetected joints, which forces us to ignore those points for calculating the correlation. The computed correlation coefficient for the 3DPW test dataset for each joint is presented in Figure~\ref{fig:correlation}. The average coefficient $\overline{r}=0.67$ indicates a strong correlation between ED and SE. This suggests that the differences in the estimated joint positions by SPIN and OpenPose models capture the error of SPIN model with respect to the ground truth.  We leverage this information and explore three techniques that use ED to estimate confidence for the recovered mesh.

\subsection{Using Raw $\text{ED}$ Values}

For a given image, $\text{ED}$ is a $K$-dimensional vector that stores the differences between joints' location estimates from SPIN and OpenPose models.  We can use these values to decide whether or not the mesh is ``good'' as follows
\begin{equation}
    y_\text{mesh} = 
    \begin{cases}
        \text{good} & \text{if } \max\ \text{ED} \leq \text{threshold} \\
        \text{bad} & \text{otherwise.}
    \end{cases}
\end{equation}
We can use a similar argument to identify the worst joint:
\begin{equation}
    k_\text{worst} = \underset{k}{\arg\max}\ \text{ED}. 
\end{equation}

\begin{figure}
    \setlength{\tabcolsep}{1pt} 
  \centering
  \begin{tabular}{c c c c c}
        {} & \multicolumn{2}{c}{Without Occlusion} & \multicolumn{2}{c}{With Occlusion} \\
        \rotatebox{90}{\hspace{0.55cm}Input} &
        \includegraphics[width=0.23\linewidth]{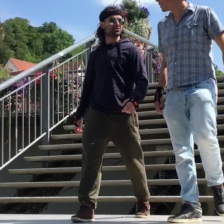} & 
        \includegraphics[width=0.23\linewidth]{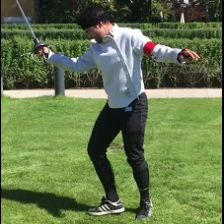} & 
        \includegraphics[width=0.23\linewidth]{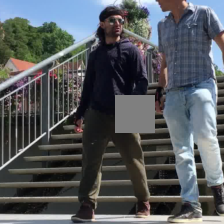} & 
        \includegraphics[width=0.23\linewidth]{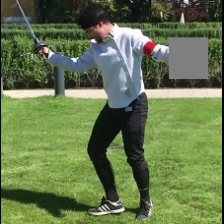} \\
        \rotatebox{90}{\hspace{0.55cm}Output} &
        \includegraphics[width=0.23\linewidth]{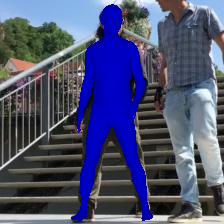} & 
        \includegraphics[width=0.23\linewidth]{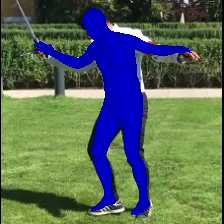} & 
        \includegraphics[width=0.23\linewidth]{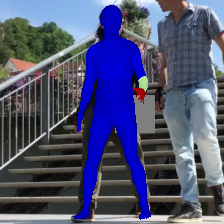} & 
        \includegraphics[width=0.23\linewidth]{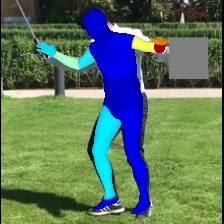} \\
  \end{tabular}
    \caption{The error distribution on the estimated mesh changes when part of the human is occluded.  For example, when a squared occluder is pasted onto the left hand, the model successfully identifies that is the least reliable region of the mesh (red regions on the mesh).}
    \label{fig:occlusionEffect}
\end{figure}

\subsection{Using Linear Regression}

Plots shown in Figure~\ref{fig:correlation} suggest a positive correlation between $\text{SE}$ and $\text{ED}$ (for all $14$ joints), which suggests that it is possible to estimate $\text{SE}$ given $\text{ED}$ for a given joint.  We are interested in estimating $\text{SE}$, since it represents the true SPIN error as computed using ground truth data.  We do not have ground truth data at inference time, so instead we estimate $\text{SE}$ using $\text{ED}$, which we can easily compute using SPIN and OpenPose models.  Therefore, we fit a linear regressor 
\begin{equation}
    \text{SE}_{k} = (m_{k}) ({\text{ED}_k})+c_{k}
    \label{eq:line}    
\end{equation}
that predicts $\text{SE}_{.,k}$ given observation $\text{ED}_{.,k}$, where $k \in [1,K]$.  Given a new image, 1) compute $\text{ED}$, 2) use the trained linear regressor in Eq.~\ref{eq:line} to estimate $\text{SE} \in \mathtt{R}^K$, and 3) use the estimated $\text{SE}$ to decide whether or not mesh is ``good'' or to identify ``good'' and ``bad'' joints using the approach discussed in the previous section.  Just substitute $\text{SE}$ in place of $\text{ED}$.

\begin{table*}
  \centering
  \begin{tabular}{|c|c|c|c|c|c|c|c|c|}
    \hline
        Dataset & \multicolumn{2}{|c|}{PCC} & \multicolumn{2}{|c|}{Mesh} & \multicolumn{2}{|c|}{WJ-R1} & \multicolumn{2}{|c|}{WJ-R3} \\
    \hline
        Regular/Occluded & R & O & R & O & R & O & R & O \\
    \hline
        3DPW & 0.67	& 0.735	& 79.2\% & 86.2\% & 42.2\% & 45.4\%	& 70.6\% & 73.8\% \\
    \hline
        3DOH & 0.665 & 0.707 & 81.6\% & 88\% & 37.3\% & 40.5\% & 64.2\% & 66.8\% \\
    \hline
        H36M-P1 & 0.492	& 0.545 & 71.9\% & 82.1\% & 42.4\% & 43.9\% & 76.4\% & 75.1\% \\
    \hline
  \end{tabular}
  \caption{Model Evaluation. Pearson Coefficient Correlation (PCC), model accuracy in separating accurate and faulty meshes (Mesh), and model performance on detecting the least reliable joints, i.e., worst joints (WJ), are presented in this table.  Model is allowed a single guess for Rank 1 (R1) and it is allowed three guesses for Rank 3 (R3).}
  \label{tab:result}
\end{table*}

\subsection{Classifiers}\label{classifier}

The previous two approaches of using $\text{ED}$ to classify recovered human body meshes and joints treat each joint separately.  We now propose an approach that looks at all $K$ joints simultaneously to classify the mesh and identify the worst joint.  Specifically, we use two multi-linear perceptron networks that use $\text{ED}$ to classify mesh and identify the worst joint, respectively.

The Mesh Classifier (MC) network is a binary classifier containing three hidden linear layers that contain $10$, $8$, and $6$ neurons respectively with ReLU activation functions.  Input to MC is $\text{ED}$ and it outputs whether or not the recovered mesh is reliable, i.e., all parts of the human body are visible in the image.  MC network is trained using binary cross-entropy.  The ground truth data for training MC is constructed using $\text{SE}$ scores---if $\text{SE}_{.,k} \ge \text{threshold}$ for any $k$ then the mesh is deemed unreliable, where $\text{SE}_{.,k}$ is the $\text{SE}$ score for joint $k$.  Under this regime 
\begin{equation}
    y_\text{mesh} = f_\text{MC} ( \text{ED} ).
\end{equation}

The Worst Joint Classifier (WJC) network is a $K$-class classification network.  It comprises three hidden layers containing $28$, $56$ and $28$ neurons, respectively.  Hidden layers use ReLU activation functions.  $\text{ED}$ is fed into WJC, and WJC is trained using cross-entropy.  The ground truth data for training  WJC is constructed from $\text{SE}$.  We simply encode $\text{SE}$ using one-hot-encoded form with $1$ at $\underset{k}{\arg\max}\ \text{SE}$ and $0$ elsewhere.  Using WJC, 
\begin{equation}
    k_\text{worst} = f_\text{WJC}( \text{ED} ).
\end{equation}

\section{Experiments and Results}\label{experiment}

We use 3DPW~\cite{von2018recovering} and Human3.6M~\cite{ionescu2013human3} (S9 and S11) datasets for model training and testing. In addition, we use 3DOH~\cite{zhang2020object} dataset for testing only.  The threshold used in Section~\ref{method} is set at $10$ mm, i.e., if the difference between an estimated joint location and the ground truth joint location is higher than $10$ mm, the mesh recovered by the SPIN model is labelled inaccurate.  We also created occluded versions of the three datasets where a randomly selected joint is occluded using a square occluder in each image. 

Figure~\ref{fig:correlation} (rows 1 and 3) shows scatter plots of $\text{SE}$ vs $\text{ED}$ for every joint for the unoccluded 3DPW dataset.  These plots also show Pearson correlation coefficient for each joint, which suggests that $\text{ED}$ is positively correlated with $\text{SE}$. This is good news, since it suggests that in the absence of $\text{SE}$, which is not available at inference time, we can use $\text{ED}$ to compute an error estimate for the recovered mesh.  Consider the $\text{ED}$ vs. $\text{SE}$ plot for right-wrist joint in Figure~\ref{fig:correlation} (first row, right most figure).  The plot identifies four regions labelled (a), (b), (c) and (d).  Points in the regions (a) and (d) have a negative effect on the correlation. Points in region (a) suggest that there are several situations where both models are inaccurate, but that they agree with each other.  Thus, we conclude that when the OpenPose and SPIN estimates are close to each other, it does not necessarily mean that the recovered human mesh is accurate.  Rather, it may be that joint estimates from both models are close to each other but far from the ground truth locations. Figure~\ref{fig:failCases} (input/output pair on the left) depicts such a case.  Here both models are in agreement with each other, however, both models fail to detect the right wrist due to self-occlusion and the presence of other people.  Points in region (d) represent cases where although the estimated values of SPIN and OpenPose model are different, the SPIN model is accurate. In other words, in some cases, a measurable difference in OpenPose and SPIN outputs does not indicate an inaccurate mesh reconstruction by the SPIN model. The right input/output pair in Figure~\ref{fig:failCases} shows an example of such a case. The SPIN model is successful in estimating the right wrist of the person, however, OpenPose model makes a mistake and selects the other person's hand position as the correct location for the right wrist.  Despite the points in regions (a) and (d), the average Pearson correlation coefficient for all joints is $\overline{r}=0.67$, indicating a strong correlation between $\text{ED}$ and $\text{SE}$ for all the joints.  This confirms our intuition that $\text{ED}$ is a good proxy for $\text{SE}$.

We performed a similar analysis as shown in Figure~\ref{fig:correlation} (rows 2 and 4) for occluded dataset, where a square occluder is pasted on a randomly selected joint.  The average Pearson correlation coefficient obtained under these settings is $\overline{r}=0.735$, which is even higher than the value computed for the unonccluded case.  This suggests two things: 1) that the proposed model is robust to occlusions and 2) $\text{ED}$ is even more positively correlated with $\text{SE}$.  Table~\ref{tab:result} shows the Pearson correlation coefficient for different test datasets, and it shows Pearson correlation coefficient is higher for occluded datasets.  In addition, 3DPW and 3DOH datasets have higher coefficient values since these exhibit higher occlusion levels.  Figure~\ref{fig:occlusionEffect} illustrates two instances of the model's behavior towards occlusion.  Our model predicts that the recovered mesh is correct when there are no occlusions, however, the model correctly identifies the left wrist region of the recovered mesh to be unreliable when a square occluder is used to hide this joint in the input image.

\begin{figure}
    \setlength{\tabcolsep}{1pt} 
  \centering
  \begin{tabular}{c c c c c}
        Input & MC & MC-GT & WJC & WJC-GT \\
        \includegraphics[width=0.19\linewidth]{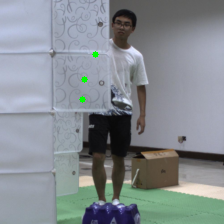} & 
        \includegraphics[width=0.19\linewidth]{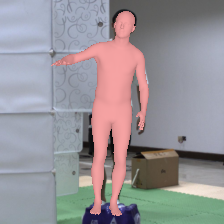} & 
        \includegraphics[width=0.19\linewidth]{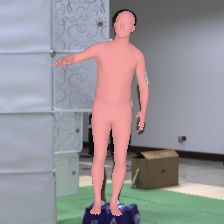} & 
        \includegraphics[width=0.19\linewidth]{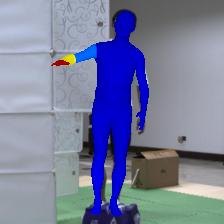} & 
        \includegraphics[width=0.19\linewidth]{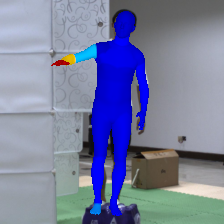} \\
        \includegraphics[width=0.19\linewidth]{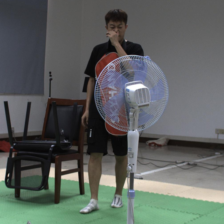} & 
        \includegraphics[width=0.19\linewidth]{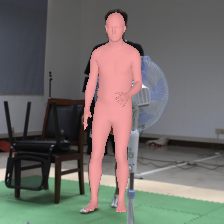} & 
        \includegraphics[width=0.19\linewidth]{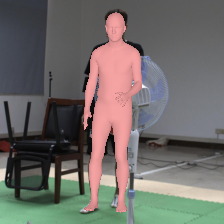} & 
        \includegraphics[width=0.19\linewidth]{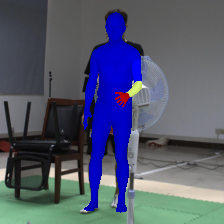} & 
        \includegraphics[width=0.19\linewidth]{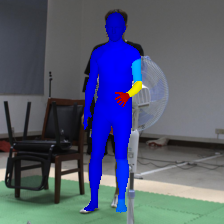} \\
        \includegraphics[width=0.19\linewidth]{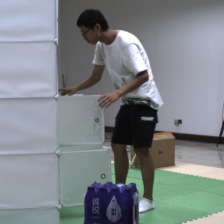} & 
        \includegraphics[width=0.19\linewidth]{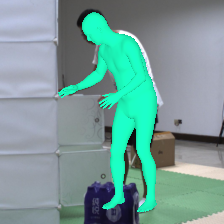} & 
        \includegraphics[width=0.19\linewidth]{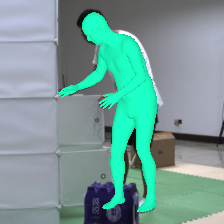} & 
        \includegraphics[width=0.19\linewidth]{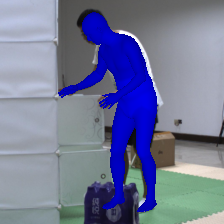} & 
        \includegraphics[width=0.19\linewidth]{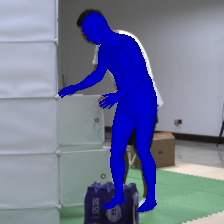} \\
        \includegraphics[width=0.19\linewidth]{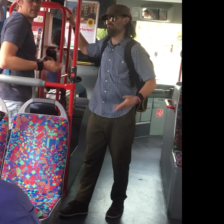} & 
        \includegraphics[width=0.19\linewidth]{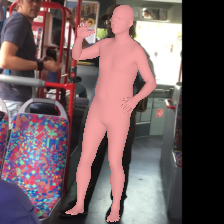} & 
        \includegraphics[width=0.19\linewidth]{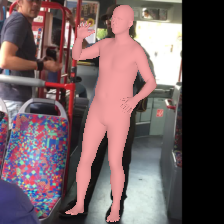} & 
        \includegraphics[width=0.19\linewidth]{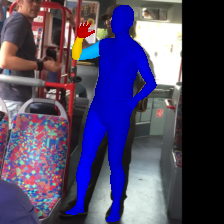} & 
        \includegraphics[width=0.19\linewidth]{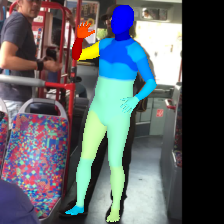} \\
        \includegraphics[width=0.19\linewidth]{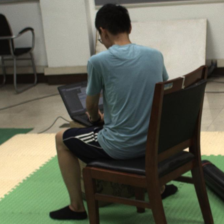} & 
        \includegraphics[width=0.19\linewidth]{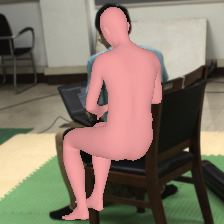} & 
        \includegraphics[width=0.19\linewidth]{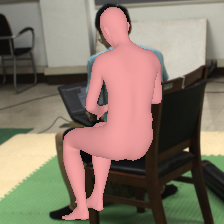} & 
        \includegraphics[width=0.19\linewidth]{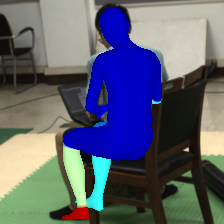} & 
        \includegraphics[width=0.19\linewidth]{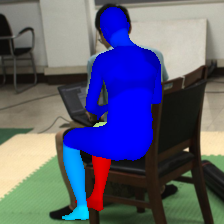} \\
        \includegraphics[width=0.19\linewidth]{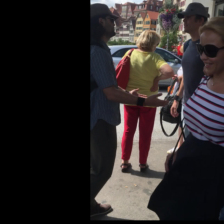} & 
        \includegraphics[width=0.19\linewidth]{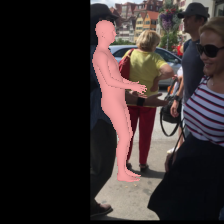} & 
        \includegraphics[width=0.19\linewidth]{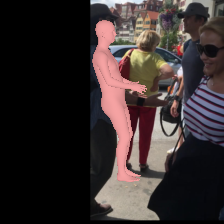} & 
        \includegraphics[width=0.19\linewidth]{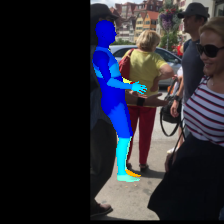} & 
        \includegraphics[width=0.19\linewidth]{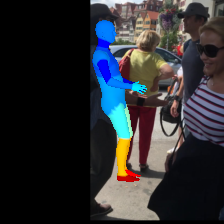} \\
  \end{tabular}
    \caption{Qualitative Results.  Input images are shown in the left column.  The next two columns contain the mesh classifier output and the ground truth.  Unreliable meshes are shown in light pink.  The fourth column highlights the least reliable joints.  Red regions on the mesh correspond to the least reliable joints.  The last column shows the ground truth for the least reliable joints.}
    \label{fig:qualitative}
\end{figure}

\begin{table}
  \centering
  \begin{tabular}{|c|c|c|c|c|c|}
    \hline
        \multicolumn{2}{|c|}{Datasets} & Metric & $\text{ED}$ & L. Regressor & Classifier\\
    \hline
        \multirow{6}{*}{\rotatebox{90}{3DPW}} & \multirow{3}{*}{R} & Mesh & 71.2 & 75.3 & 79.2\\
    \cline{3-6}
                           &                    & WJ-R1 & 27.8 & 38.2 & 42.2\\
    \cline{3-6}
                           &                    & WJ-R3 & 61.5 & 68.8 & 70.6\\
    \cline{2-6}
                           & \multirow{3}{*}{O} & Mesh & 82.7 & 81.2 & 86.2\\
    \cline{3-6}
                           &                    & WJ-R1 & 30.7 & 42.17 & 45.4\\
    \cline{3-6}
                           &                    & WJ-R3 & 65.5 & 72.2 & 73.8\\
    \hline
        \multirow{6}{*}{\rotatebox{90}{3DOH}} & \multirow{3}{*}{R} & Mesh & 80.8 & 82.9 & 81.6\\
    \cline{3-6}
                           &                    & WJ-R1 & 22 & 30.4 & 37.3\\
    \cline{3-6}
                           &                    & WJ-R3 & 54.5 & 70.2 & 64.2\\
    \cline{2-6}
                           & \multirow{3}{*}{O} & Mesh & 88.6 & 88 & 88\\
    \cline{3-6}
                           &                    & WJ-R1 & 22.5 & 31.7 & 40.5\\
    \cline{3-6}
                           &                    & WJ-R3 & 55.2 & 67.5 & 66.8\\
    \hline
        \multirow{6}{*}{\rotatebox{90}{H36M-P1}} & \multirow{3}{*}{R} & Mesh & 66.5 & 67.8 & 71.9\\
    \cline{3-6}
                           &                    & WJ-R1 & 17.8 & 29 & 42.4\\
    \cline{3-6}
                           &                    & WJ-R3 & 58.6 & 66.5 & 76.4\\
    \cline{2-6}
                           & \multirow{3}{*}{O} & Mesh & 79.9 & 78.2 & 82.1\\
    \cline{3-6}
                           &                    & WJ-R1 & 22.9 & 36 & 43.9\\
    \cline{3-6}
                           &                    & WJ-R3 & 64.1 & 69.5 & 75.1\\
    \hline
  \end{tabular}
  \caption{Ablation Study.  Comparing the method that uses raw $\text{ED}$ values (column 3), linear regressor (column 4), and classifier based method (column 5) for classifying unreliable meshes and identifying the least reliable joints.  Mesh refers to mesh reliability classification results, WJ-R1 refers to the results for identifying the worst joint (least reliable) when a single guess is allowed, and WJ-R3 refers to results for identifying the worst joint in three guesses.}
  \label{tab:ablation}
\end{table}

We exploit the positive correlation between $\text{ED}$ and $\text{SE}$ to estimate the error in the human body mesh recovered by SPIN.  The proposed method also highlights the least reliable region of the recovered mesh.  Table~\ref{tab:result} lists our model's performance at identifying an inaccurate mesh.  Additionally, this table also includes model's performance at identifying the least reliable joint.  There is no baseline, since, to the best of our knowledge, ours is the first attempt at performing error estimation for single-image human body mesh reconstruction scenarios.  For example, while the model was never trained on 3DOH dataset, it is able to identify an inaccurate mesh with $88$\% accuracy.  The model is also able to identify the least reliable joint $40.5$\% accuracy.  This number jumps to $66.8$\% when the model is allowed three guesses for the least reliable joint.  These numbers are considerably higher than randomly selecting the least reliable joint.  A similar trend is visible for 3DPW and H36M-P1 datasets.   

\begin{figure*}
    \setlength{\tabcolsep}{1pt} 
  \centering
  \begin{tabular}{c c c c c c c c c c c}
        \rotatebox{90}{\hspace{0.45cm}Input} &
        \includegraphics[width=0.08\linewidth]{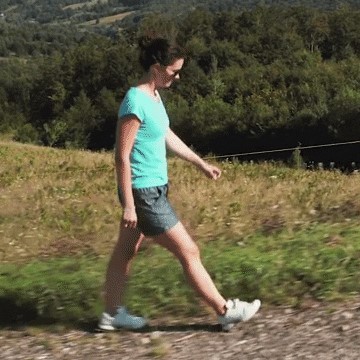} & 
        \includegraphics[width=0.08\linewidth]{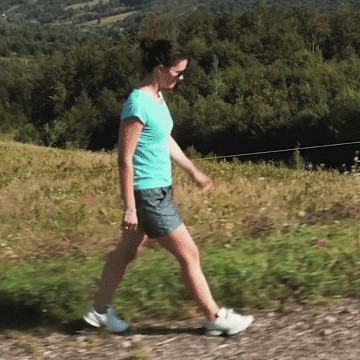} & 
        \includegraphics[width=0.08\linewidth]{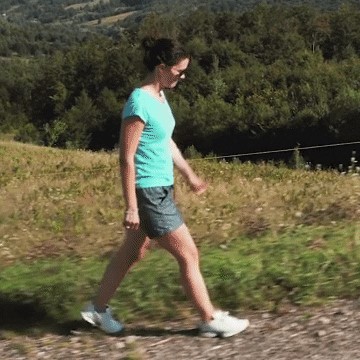} & 
        \includegraphics[width=0.08\linewidth]{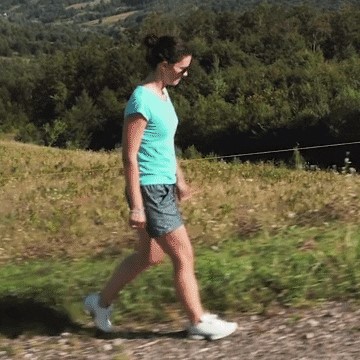} &
        \includegraphics[width=0.08\linewidth]{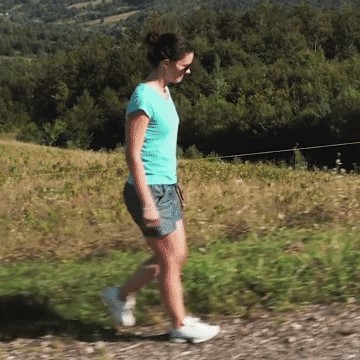} & 
        \includegraphics[width=0.08\linewidth]{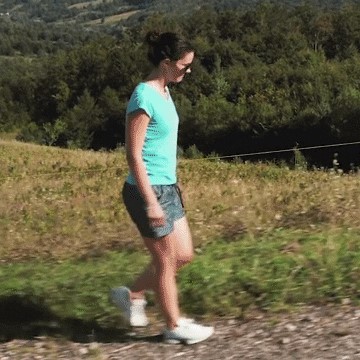} & 
        \includegraphics[width=0.08\linewidth]{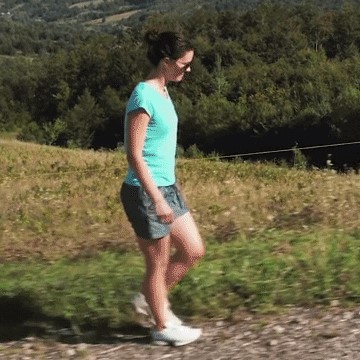} & 
        \includegraphics[width=0.08\linewidth]{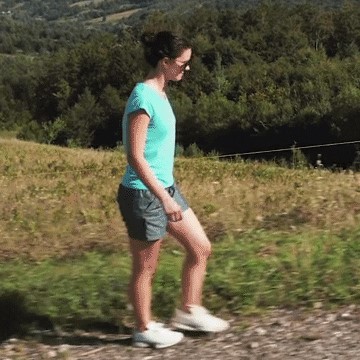} &
        \includegraphics[width=0.08\linewidth]{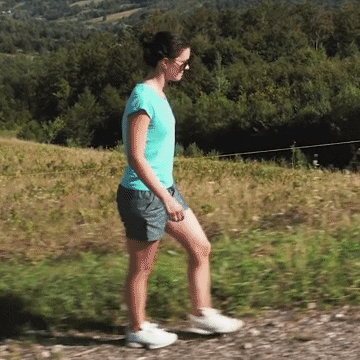} & 
        \includegraphics[width=0.08\linewidth]{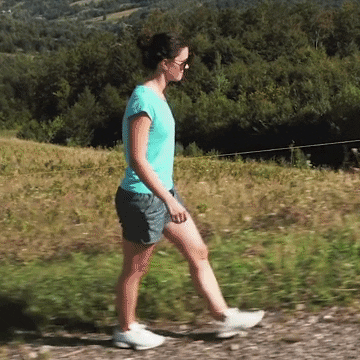} \\
        \rotatebox{90}{\hspace{0.45cm}MC} &
        \includegraphics[width=0.08\linewidth]{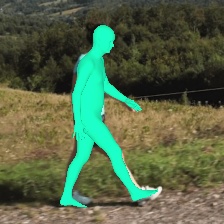} & 
        \includegraphics[width=0.08\linewidth]{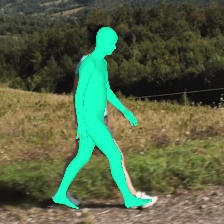} & 
        \includegraphics[width=0.08\linewidth]{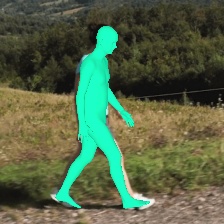} & 
        \includegraphics[width=0.08\linewidth]{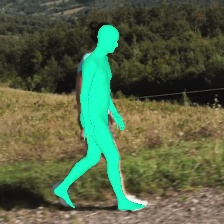} &
        \includegraphics[width=0.08\linewidth]{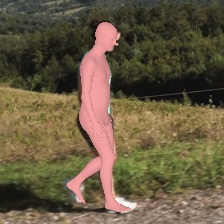} & 
        \includegraphics[width=0.08\linewidth]{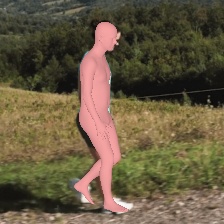} & 
        \includegraphics[width=0.08\linewidth]{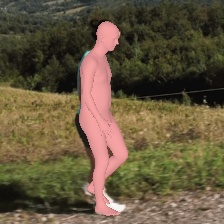} & 
        \includegraphics[width=0.08\linewidth]{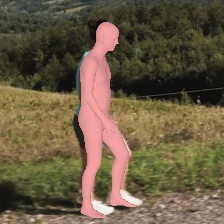} &
        \includegraphics[width=0.08\linewidth]{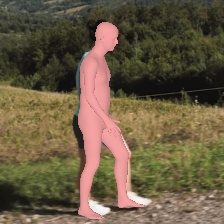} & 
        \includegraphics[width=0.08\linewidth]{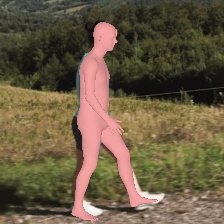} \\
        \rotatebox{90}{\hspace{0.45cm}WJC} &
        \includegraphics[width=0.08\linewidth]{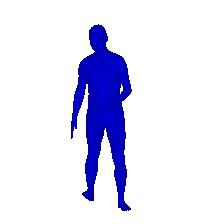} & 
        \includegraphics[width=0.08\linewidth]{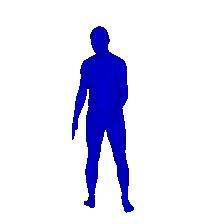} & 
        \includegraphics[width=0.08\linewidth]{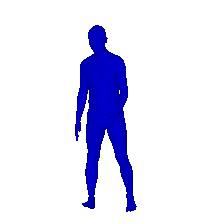} & 
        \includegraphics[width=0.08\linewidth]{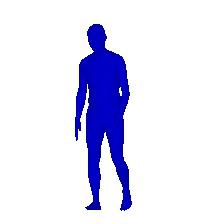} &
        \includegraphics[width=0.08\linewidth]{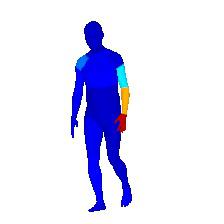} & 
        \includegraphics[width=0.08\linewidth]{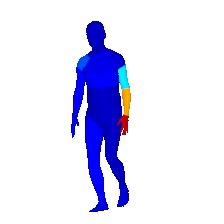} & 
        \includegraphics[width=0.08\linewidth]{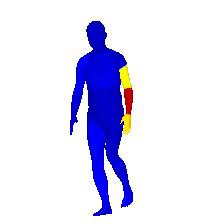} & 
        \includegraphics[width=0.08\linewidth]{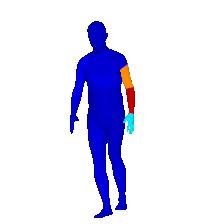} &
        \includegraphics[width=0.08\linewidth]{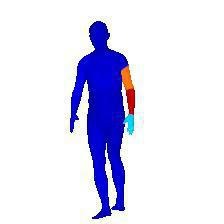} & 
        \includegraphics[width=0.08\linewidth]{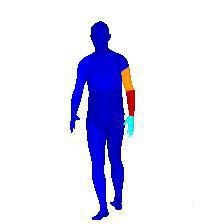} \\
        \rotatebox{90}{\hspace{0.45cm}Input} &
        \includegraphics[width=0.08\linewidth]{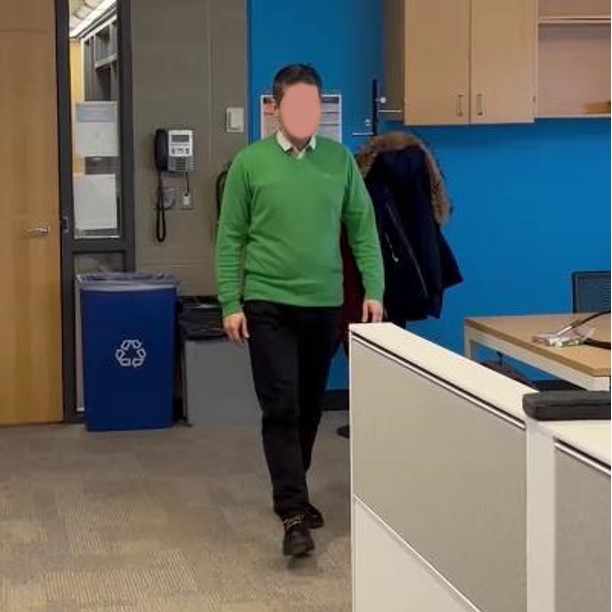} & 
        \includegraphics[width=0.08\linewidth]{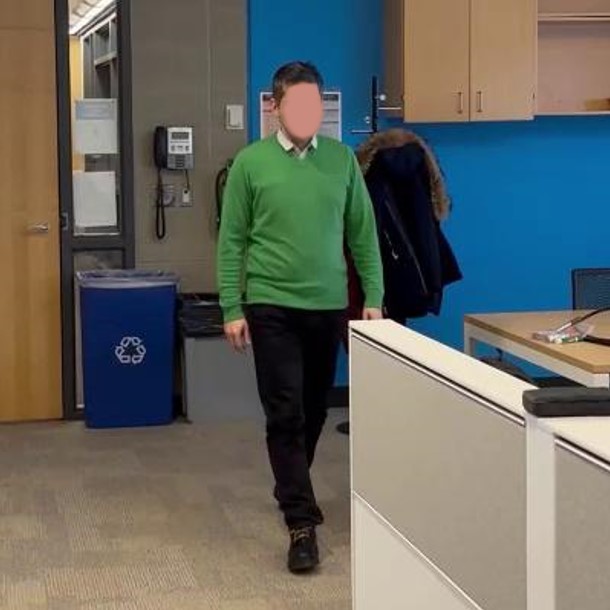} & 
        \includegraphics[width=0.08\linewidth]{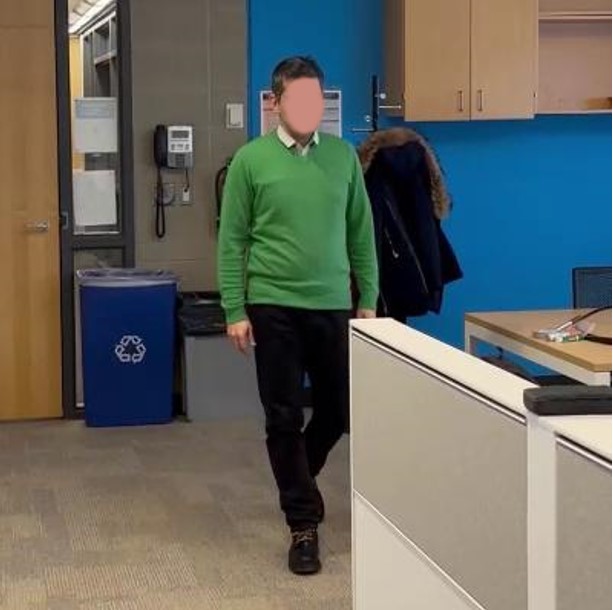} & 
        \includegraphics[width=0.08\linewidth]{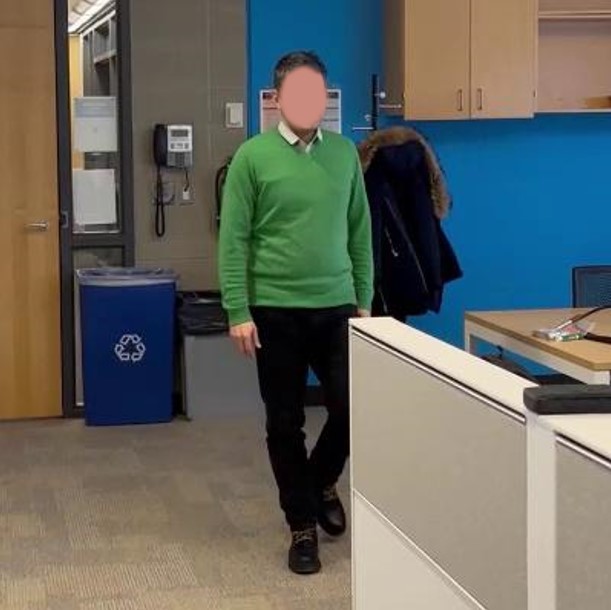} &
        \includegraphics[width=0.08\linewidth]{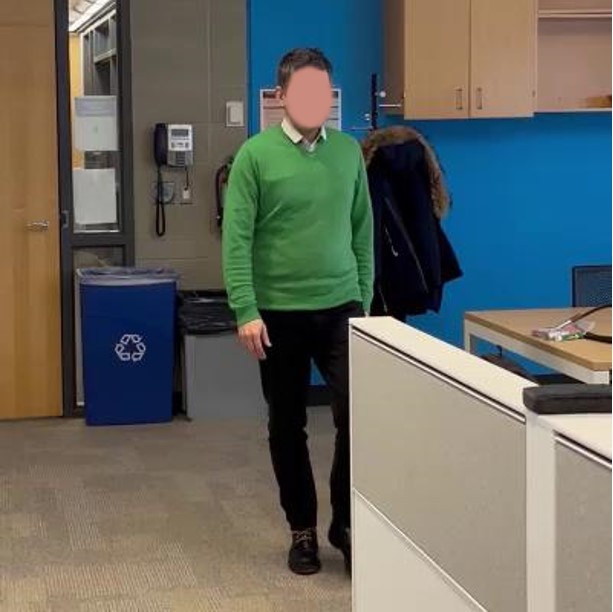} & 
        \includegraphics[width=0.08\linewidth]{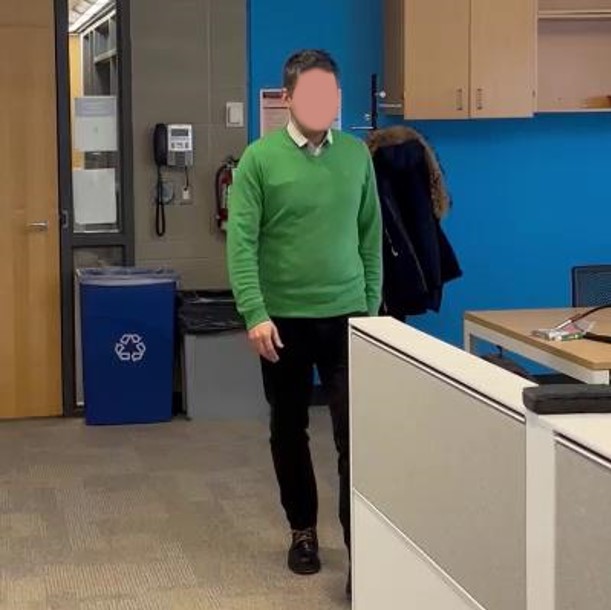} & 
        \includegraphics[width=0.08\linewidth]{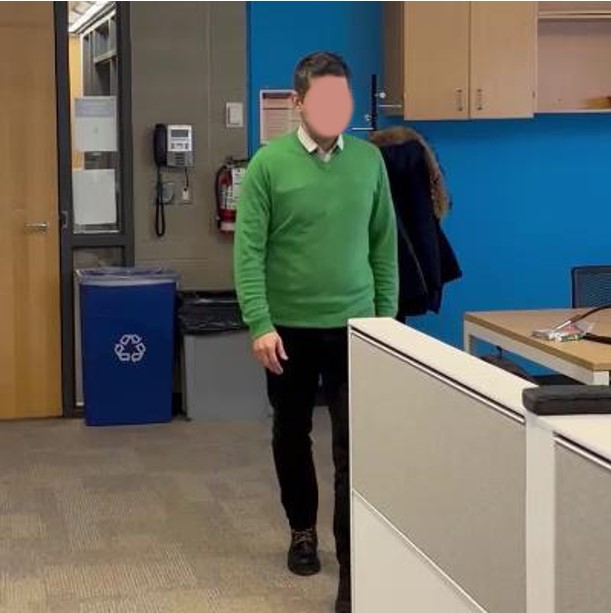} & 
        \includegraphics[width=0.08\linewidth]{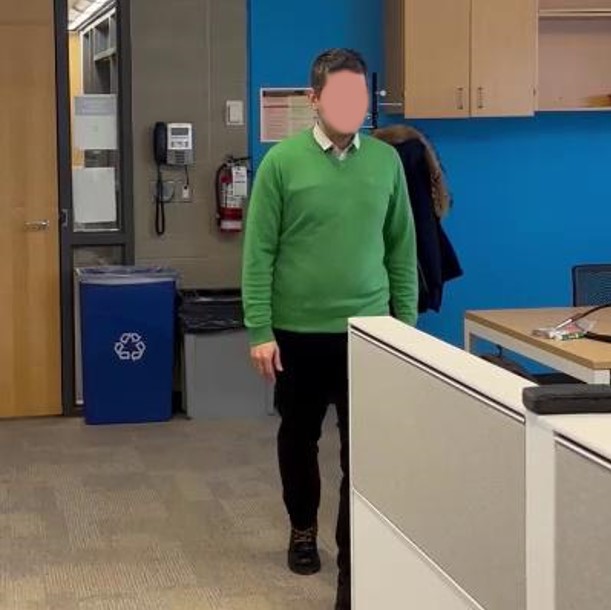} &
        \includegraphics[width=0.08\linewidth]{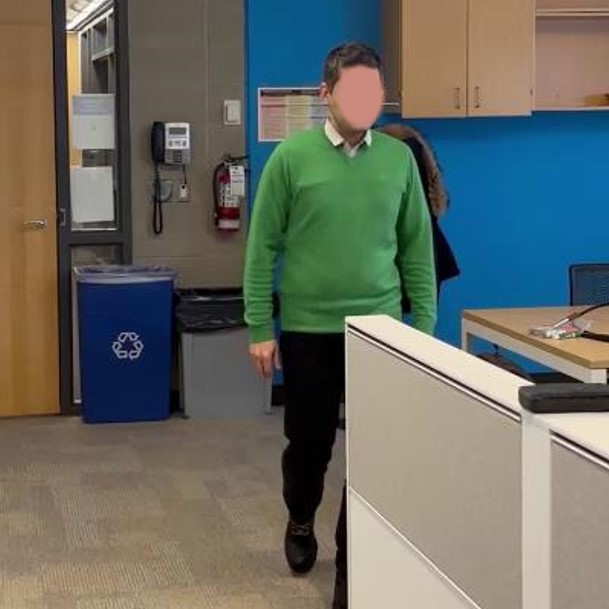} & 
        \includegraphics[width=0.08\linewidth]{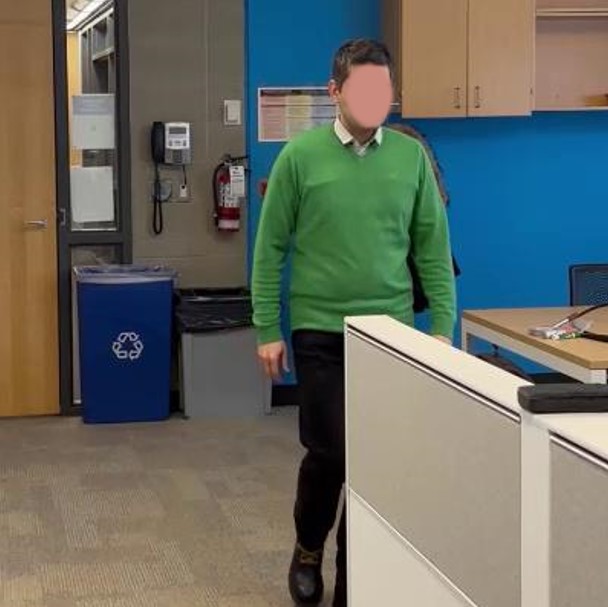} \\
        \rotatebox{90}{\hspace{0.45cm}MC} &
        \includegraphics[width=0.08\linewidth]{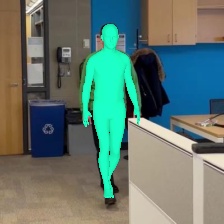} & 
        \includegraphics[width=0.08\linewidth]{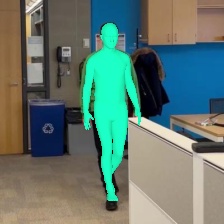} & 
        \includegraphics[width=0.08\linewidth]{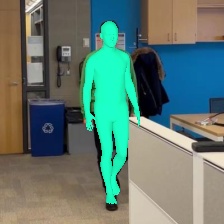} & 
        \includegraphics[width=0.08\linewidth]{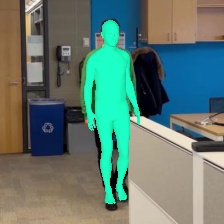} &
        \includegraphics[width=0.08\linewidth]{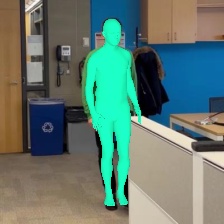} & 
        \includegraphics[width=0.08\linewidth]{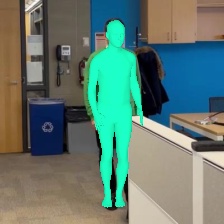} & 
        \includegraphics[width=0.08\linewidth]{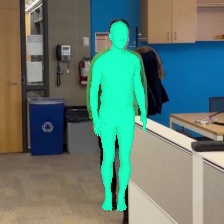} & 
        \includegraphics[width=0.08\linewidth]{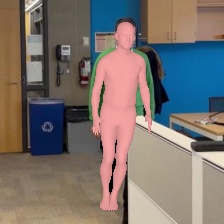} &
        \includegraphics[width=0.08\linewidth]{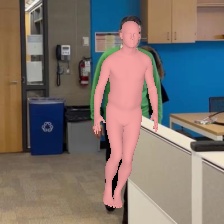} & 
        \includegraphics[width=0.08\linewidth]{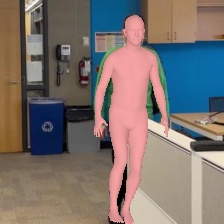} \\
        \rotatebox{90}{\hspace{0.45cm}WJC} &
        \includegraphics[width=0.08\linewidth]{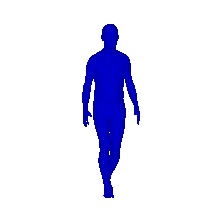} & 
        \includegraphics[width=0.08\linewidth]{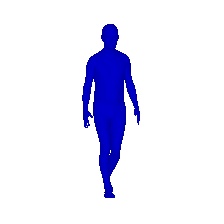} & 
        \includegraphics[width=0.08\linewidth]{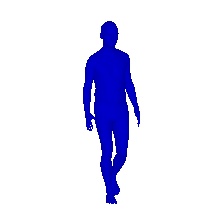} & 
        \includegraphics[width=0.08\linewidth]{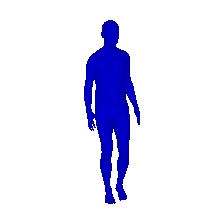} &
        \includegraphics[width=0.08\linewidth]{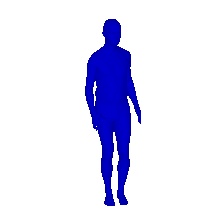} & 
        \includegraphics[width=0.08\linewidth]{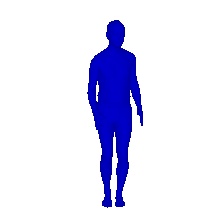} & 
        \includegraphics[width=0.08\linewidth]{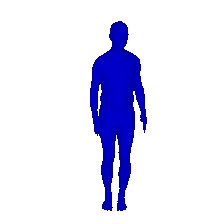} & 
        \includegraphics[width=0.08\linewidth]{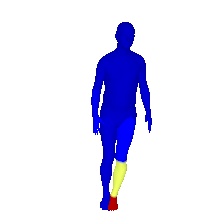} &
        \includegraphics[width=0.08\linewidth]{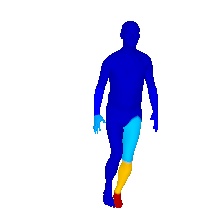} & 
        \includegraphics[width=0.08\linewidth]{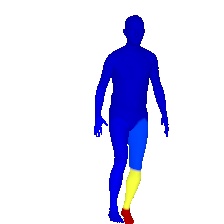} \\
  \end{tabular}
    \caption{Application on Videos.  The first row shows the input video frames.  The second row shows mesh reliability classification results.  Light pink indicates an unreliable mesh.  The last row shows least reliable joints.  Here the meshes are rotated by $90$ degrees for better visualization.  The red regions on the mesh highlight the least reliable joint.}
    \label{fig:video}
\end{figure*}

Consider Figure~\ref{fig:qualitative} that presents some qualitative results.  The first four rows show cases where the proposed model performed correctly.  Here MC denotes output from the mesh classifier and MC-GT denotes the ground truth.  WJC highlights the least reliable joint(s) and WJC-GT shows the least reliable joint ground truth.  The bottom two rows show failure cases.  Here, while the model correctly predicts that the recovered mesh is unreliable, it is unable to identify the least reliable joint correctly.  Figure~\ref{fig:video} shows an application of our method on video data. Here the top row shows input frames, the second row shows whether or not the recovered mesh is reliable, and the last row includes a visualization of the least reliable joint.  The meshes shown in the last row are rotated to better see the least reliable joints.  Our model correctly handles self-occlusions (top three rows) and occlusions due to other objects in the scene (bottom three rows).  Check the last row where the model correctly predicts that the left foot is the least reliable region of the recovered mesh since it is not visible in the image (it is occluded by the table).  The decision to decide if the recovered mesh is ``reliable'' when only left foot is not visible in the image is application specific.  For example, say a robot is simply navigating around this person then perhaps it is okay to deem the recovered mesh to be reliable.  However, if this same robot is carrying out a task that involves the left foot of this person then it is best to consider this mesh unreliable.

\subsection{Ablation Study} 
We now compare the performance of the three approaches discussed in Section~\ref{method}.  All three approaches leverage the positive correlation between $\text{ED}$ and $\text{SE}$.  Table~\ref{tab:ablation} shows the results obtained for each approach on the three datasets in both unoccluded and occluded cases. The results confirm that the classifier-based approach that combines $\text{ED}$ information from different joints outperforms the other two methods.  Method that uses raw $\text{ED}$ values posts the worst performance.  What is interesting to note is that using a classifier dramatically increases the performance of identifying the least reliable joint, both when the model is allowed a single guess and when it is allowed three guesses.  This suggests that it is beneficial to consider {\it all} joints' errors when selecting the least reliable joint.  For mesh classification, however, the improvement obtained by using a classifier-based approach over using the method that relies on raw $\text{ED}$ values is not nearly as significant.
 
\section{Conclusion}

This work develops a method for estimating the error in the human body meshes reconstructed by the SPIN model.  The model is not only able to decide whether or not a mesh is unreliable, it is also able to highlight the least reliable, i.e., having the highest error, regions on the mesh.   The proposed model uses the disagreement between joint location estimates between OpenPose and SPIN model to compute error values for the recovered mesh.  Pearson correlation coefficient studies on 3DPW dataset show this disagreement is a good proxy for the ``true'' error.  Evaluations on 3DPW, 3DPH, and H36M-P1 confirm that the model is able to estimate error in the SPIN based single-image human body mesh reconstructions in the presence of occlusions.  Furthermore, it is able to correctly estimate the error in SPIN meshes even when OpenPose estimates are incorrect.  The model is also able to identify the least reliable joints.  The ability to estimate the error in the recovered meshes is particularly important when these meshes are used in human-robot interaction scenarios. To the best of our knowledge, ours is the first method to estimate error in single-image 3D human body mesh reconstruction.

{\small
\bibliographystyle{ieee_fullname}
\bibliography{egbib}
}

\end{document}